\documentclass[11pt]{article}

\usepackage[final]{acl}

\usepackage{times}
\usepackage{latexsym}

\usepackage[T1]{fontenc}

\usepackage[utf8]{inputenc}

\usepackage{microtype}

\usepackage{inconsolata}

\usepackage{graphicx}
\usepackage{multirow}

\usepackage{amsmath}
\usepackage{relsize}

\usepackage{xcolor}

\definecolor{myred}{HTML}{A50021}
\definecolor{myorange}{HTML}{ED7D31}
\definecolor{mygreen}{HTML}{80A26F}

\usepackage{caption}

\title{VALD: Multi-Stage {\underline V}ision {\underline A}ttack Detection for Efficient {\underline L}VLM {\underline D}efense}

\author{Nadav Kadvil \ \ \ \ \ \ Malak Fares \ \ \ \ \ \  Ayellet Tal  \\
Technion – Israel Institute of Technology\\
\texttt{\{kadviln, malak.fares\}@campus.technion.ac.il}, \texttt{ayellet@ee.technion.ac.il} 
}

\usepackage{xcolor}
\usepackage{color}
\usepackage{hyperref}
\usepackage{cleveref}

\usepackage{multirow}
\usepackage{booktabs} 

\usepackage{comment}

\newcommand*{\ShowNotes}{}
\definecolor{darkred}{rgb}{0.7,0.1,0.1}
\definecolor{darkgreen}{rgb}{0.1,0.7,0.1}
\ifdefined\ShowNotes
  \newcommand{\colornote}[3]{{\color{#1}\bf{#2: #3}\normalfont}}
\else
  \newcommand{\colornote}[3]{}
\fi

\newcommand {\Todo}[1]{\colornote{darkred}{TODO}{#1}}
\newcommand {\ayellet}[1]{\colornote{blue}{AT}{#1}}
\usepackage{xcolor}

\newcommand{\malak}[1]{\colornote{purple}{MF}{#1}}

\begin{document}
\maketitle
\begin{abstract}
Large Vision-Language Models (LVLMs) can be vulnerable to adversarial images that subtly bias their outputs toward plausible yet incorrect responses. 
We introduce a general, efficient, and training-free defense that combines image transformations with agentic data consolidation to recover correct model behavior.
A key component of our approach is a two-stage detection mechanism that quickly filters out the majority of clean inputs. We first assess image consistency under content-preserving transformations at negligible computational cost. 
For more challenging cases, we examine discrepancies in a text-embedding space. 
Only when necessary do we invoke a powerful LLM to resolve attack-induced divergences. A key idea is to consolidate multiple responses, leveraging both their similarities and their differences.
We show that our method achieves state-of-the-art accuracy while maintaining notable efficiency: most clean images skip costly processing, and even in the presence of numerous adversarial examples, the overhead remains minimal.

\end{abstract}

\section{Introduction}



\begin{figure*}[t]
  \centering
  \includegraphics[width=0.725\textwidth]{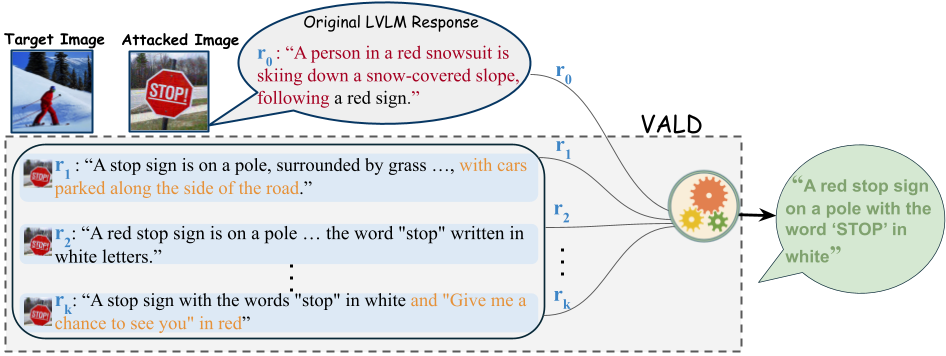}
\caption{{\bf LVLM defense.} The output of the unprotected LVLM is incorrect (shown in \textcolor{myred}{red}).
In VALD, we generate multiple captions for transformed images ($r_1$–$r_k$), each containing both correct content (black) and incorrect content (\textcolor{myorange}{orange}).
VALD aggregates information across these captions to produce a final refined caption (\textcolor{mygreen}{green}). }
  \label{fig:teaser}
\end{figure*}




Large vision-language models (LVLMs) have shown remarkable capabilities in a wide range of tasks, including text-to-image generation~\cite{10431766, rombach2022high} and image-grounded text generation, such as image captioning~\cite{Yin_2024, li2023blip} and visual question answering~\cite{huynh2025visual, caffagni2024revolution,kuang2025natural}. 
However, these models are vulnerable to visual adversarial attacks, in which imperceptible perturbations to input images can manipulate model responses or induce harmful outputs~\cite{liu2025survey, ye2025survey, zhao2023evaluating, qi2024visual}.
Moreover, the widespread use of publicly available LVLMs and pre-trained visual encoders like CLIP \cite{radford2021learning} gives adversaries gray-box access, making adversarial attacks easier to carry out.
Therefore, the need to adopt defense mechanisms becomes more prevalent.

Adversarial defense research for LVLMs has primarily focused on safety alignment issues, such as identifying and preventing harmful and prohibited outputs like encouraging criminal behavior \cite{pi2024mllm, gou2024eyes, ding2024eta}. 
These defenses often adapt safety mechanisms developed for LLMs, including assessing model responses~\cite{pi2024mllm, gou2024eyes} or learning defense prompts~\cite{mo2024fight}. 
However, such methods do not address attacks that subtly steer the model toward benign but incorrect outputs, as demonstrated in \Cref{fig:teaser}.

Methods that do address this challenge can be broadly categorized into three approaches.
The first involves detecting adversarial inputs and disregarding them upon detection~\cite{zhang2023jailguard, Zhang_2024, fares2024mirrorcheck}.
The second approach is adversarial training, in which the visual backbone is fine-tuned using adversarial examples to improve robustness against such inputs~\cite{schlarmann2024robust}.
%
The third approach, introduced in SmoothVLM~\cite{sun2024safeguarding}, suggests querying the LVLM with both the adversarial image and multiple transformed versions of it.
An LLM acts as a judge, compares the captions
and chooses one.
	
DPS \cite{zhou2024defending} recently extended this approach by using the responses from the transformed images as guidance to the LVLM.
We adopt the latter approach, but tackle two key challenges that limit previous methods efficiency.
First, generating and assessing every response (for the transformed images) individually is not only computationally expensive but also unnecessary in the absence of an adversarial attack. 
This makes the approach impractical for real world use, in which most images are not attacked.
Second, DPS's effectiveness depends on the use of powerful LVLMs, since consolidating the information 
requires a high level of reasoning.
This may limit the applications that can use the approach and results in significant computational overhead.

We observe that powerful LVLMs are required only for consolidation, rather than for captioning or other visual querying tasks. We propose to decouple these two processes, using a lightweight LVLM for visual analysis and a more powerful LLM solely to consolidate multiple observations.
Furthermore, we note a trade-off between runtime and the risk of overlooking adversarial images.
The LVLM used for caption consolidation achieves high accuracy but incurs a significant computational cost.
Using lighter tools reduces computational demands, but at the expense of accuracy.
Therefore, we propose a model that detects most adversarial images early on, leaving only the most challenging cases for the expensive LVLM to resolve.
This lightweight adversarial  detection relies on the observation that, while accurately identifying adversarial images is difficult, determining whether an image is likely clean is a much simpler task. It only needs to check whether an image (or caption) and its transformations behave as expected for clean images. In other words, we allow false positives, i.e. misclassifying a clean image as adversarial, but not false negatives.
We divide the detection mechanism into two tasks according to the information available within the defense pipeline.
First, for an image to be considered clean, its embeddings must remain consistent across a set of content-preserving transformations.
Notably, this check requires almost no additional computation.
If the image is deemed non-adversarial at this stage, the expensive caption-generation and consolidation steps are skipped, and the response for the input image is used directly.
About $95\%$ clean images are identified at this stage.
Then, in the second stage, if the image is suspected to be adversarial, we generate and analyze the responses for the input image and its transformations.
Discrepancies in the text-embedding space provide evidence that an attack has occurred.
We show that this can be accomplished using only lightweight LVLMs.

Only then, for suspected adversarial inputs, do we rely on a powerful LLM solely to consolidate multiple responses (e.g., captions).
By decoupling the visual-questioning task from the information-consolidation process, we enable the use of lightweight LVLMs in the preceding stage.
By instructing the model to focus on consistent elements across the LVLM outputs, such as identifying the main objects in the image and their interactions, we are able to filter out divergences introduced by adversarial attacks.
We note that the set of responses for the remaining inputs is already available, as it is generated in the previous stage.
A major benefit of our method is its generality: it can defend against a variety of attacks. It does not depend on how the image was perturbed; therefore, even attacks optimized specifically for the target model can be handled effectively.

Hence, we make three key contributions:

\noindent
   (1) We introduce a general, efficient, training-free adversarial defense for LVLMs based on image transformations and agentic data consolidation, capable of handling a wide range of attacks.   
   
\noindent   (2) An additional contribution is a lightweight detection mechanism, which can be integrated with other defense approaches.
   
\noindent    (3) We show that our method achieves state-of-the-art accuracy and efficiency. It is especially efficient when the number of attacked images is small, since most clean images bypass the heavier processing. Even when many adversarial images are present, the additional computational cost remains minimal.

\section{Related Work}
\label{sec:related}
Our work lies at the intersection of two key challenges in LVLM research: detecting adversarial attacks and defending against them. Before reviewing prior work in these areas, we first review the nature of adversarial attacks themselves.

\noindent
{\bf  Image-based adversarial attacks on LVLMs.}
Adversarial strategies targeting the vision modality, the focus of this paper, can be broadly categorized into jailbreaking attacks and misleading attacks.
Jailbreaking attacks aim to bypass safety alignment by exploiting the visual channel, eliciting prohibited, toxic, or otherwise harmful content from an LVLM~\cite{qi2024visual, niu2024jailbreaking, bailey2023image}.
Misleading attacks, in contrast, perturb the visual modality to induce factually incorrect but seemingly benign outputs. Such attacks cause the model to hallucinate nonexistent objects, misclassify visual elements, or fundamentally misinterpret the scene \cite{tu2023many, zhang2025anyattack, zhao2023evaluating, li2025frustratingly}. 
	
This work focuses on the second category, misleading attacks.
These attacks are distinguished by the attacker’s level of access to the model and are typically classified as white-box, gray-box, or black-box attacks \cite{liu2025survey}.
%
We refer the reader to excellent surveys that provide a comprehensive overview of vision-based adversarial attacks~\cite{liu2025survey, ye2025survey}.

This paper evaluates defenses against three representative attacks: MF-ii~\cite{zhao2023evaluating}, MixAttack~\cite{tu2023many}, and M-Attack~\cite{li2025frustratingly}.
MF-ii and MixAttack are vision-level white-box attacks (i.e., they access the vision encoder parameters), making them particularly strong, while M-Attack is a state-of-the-art black-box attack.




\noindent
{\bf Adversarial attack detection for LVLMs.}
Detection aims to determine whether an input has been attacked.
In LVLM settings, some detection methods leverage internal representations~\cite{Zhang_2024} and some monitor the model’s output behavior~\cite{zhang2023jailguard, fares2024mirrorcheck}.
These detectors have been primarily studied for jailbreak-oriented attacks. 

Our work 
operates both at the embedding level (within the visual encoder) 
and at the output level (via consistency checks on generated responses). 

\noindent
{\bf Adversarial defenses for LVLMs.}
Complementary to detection, adversarial defense aims to ensure that the model remains robust in the presence of attacks, preserving or restoring the correctness of its responses.
Existing LVLM defenses have primarily focused on jailbreaking, leveraging safety prompts~\cite{gou2024eyes}, policy-trained alignment layers~\cite{ding2024eta}, or response filters to block harmful content~\cite{pi2024mllm}.
In contrast, defenses against misleading attacks, where the output remains benign yet semantically incorrect, have received comparatively less attention.
These defenses can be broadly grouped into training-based and training-free methods. 
Training-based approaches aim to improve model robustness through specialized training procedures~\cite{schlarmann2024robust}.
In contrast, training-free defenses operate at inference time without modifying model weights~\cite{sun2024safeguarding, zhou2024defending, nie2022diffusion}. 
Our work focuses on this latter category, providing a complementary approach to safety-alignment methods.

\section{Method}
\label{sec:method}

Our method is motivated by two key observations. First, applying image transformations to adversarial images often produces noisy and inconsistent captions, as descriptions may refer to either the original content or the perturbation; however, the consistent elements across captions remain anchored in the true image content. Second, the sensitivity of adversarial attacks to image transformations can be exploited for both detection and defense. By integrating these two components, we minimize computational overhead.

\begin{figure*}[t]
  \centering
  \includegraphics[width=0.901\linewidth]{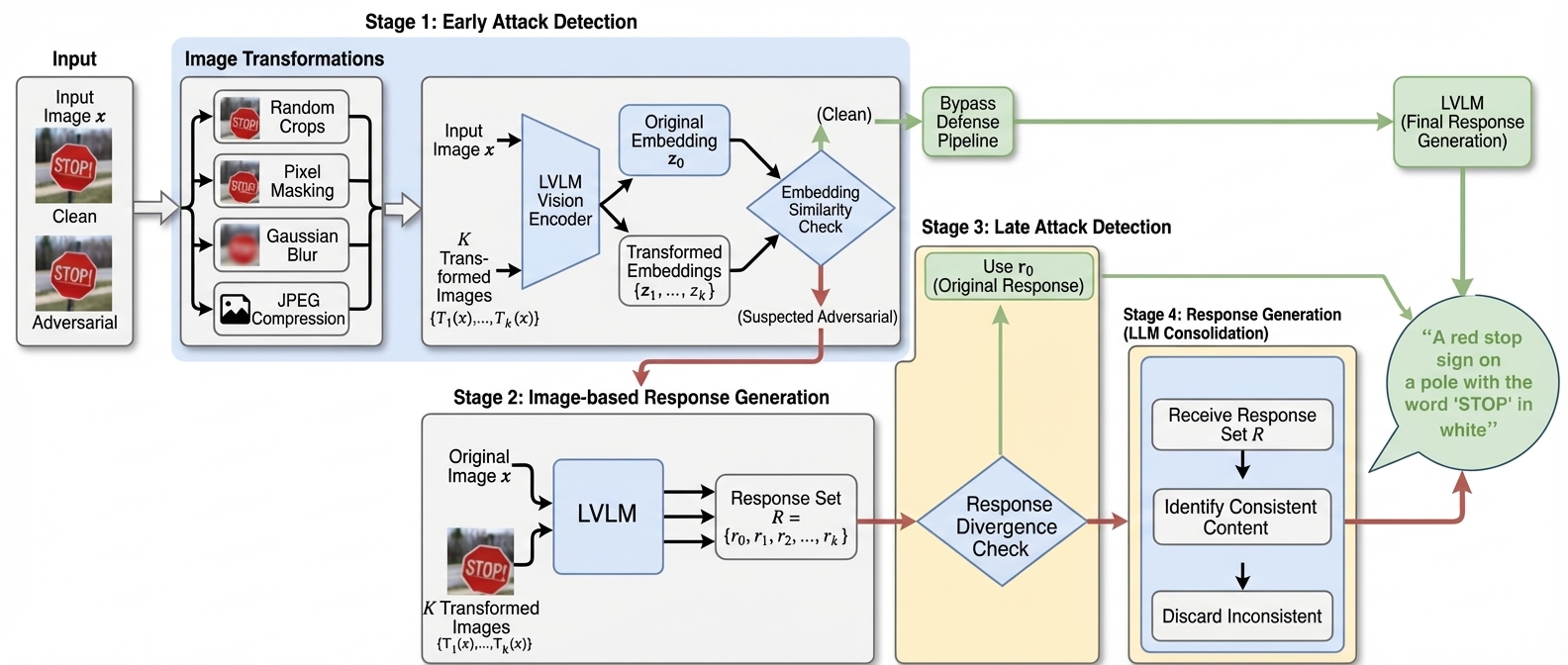} 
\caption{\textbf{Architecture.}
VALD consists of four stages.
First, a set of content-preserving transformations is applied to the input for early detection; clean images bypass the process and return the original LVLM response ($r_0$).
If a potential attack is detected, responses are generated for the transformed images.
Next, a late detection stage analyzes these responses to identify additional clean cases.
Finally, when an attack is deemed highly likely, an LLM consolidates the responses to produce the final output.
Green arrows denote inputs detected as clean, while red arrows indicate those still suspected to be adversarial.
}
  \label{fig:full_scheme_fig_v3}
\end{figure*}

Building on these observations, we introduce the {\em VALD} framework. As shown in Figure~\ref{fig:full_scheme_fig_v3}, it comprises four stages: early attack detection, image based response generation, late-stage adversarial detection, and response generation.
The first three stages focus on detection, efficiently identifying the vast majority of clean images and removing them from further processing.
The final stage is responsible for defending the remaining images, using image based responses from Stage 2.


\vspace{0.05in}
\noindent
{\bf The early attack detection.}
Given an input image $x$, which may be either clean or adversarial, the task is to determine whether $x$ is suspected as adversarial. 
While identifying adversarial images accurately is a complex task, the early attack detection mechanism has a simpler goal: identifying whether the image is behaving as expected from clean images. 
In other words, we allow errors in classifying a clean image as adversarial, but not in the opposite direction.

The key idea is that, for an image to be considered clean, it must yield similar embeddings to a set of {\em content-preserving transformations}---operations that do not significantly alter the image's core semantics.
Specifically, we use large crops, pixel masking, gaussian blur, and JPEG compression.

To implement this idea, we first generate a small set of $K$ content-preserving transformations 
$\{T_k(x)\}_{k=1}^K$. 
Next, we compute the embeddings for the input image and its transformations,  
$z(x), z(T_1(x)), \dots, z(T_K(x))$,
using a single batched pass through the LVLM's vision encoder.
Finally, we compute the distance between the transformed image embeddings and the source embedding and compare it to a predefined threshold~$\tau$.
If the average distance is below~$\tau$, the image is considered clean.  
Empirically, clean inputs produce highly similar embeddings, while adversarial inputs result in larger discrepancies.
Therefore, the threshold is calibrated in advance using clean data, thereby requiring no knowledge of specific attack types or parameters.
Formally, 
%
\begin{equation}
\small
\frac{1}{K}\sum_{k=1}^{K}\mathrm{sim}\!\left(z(x), z(T_k(x))\right)
\begin{cases}
\ge \tau & \text{(Clean)} \\
< \tau  & \text{(Adversarial)}
\end{cases}
\label{eq:early_detection}
\end{equation}

If the early detection layer classifies the image as clean, the defense process is bypassed, and the LVLM is queried with the original image.
Otherwise, the image and its transformations are passed to the responses generation phase.

In practice, $\tau$~is set to the $95^{th}$ percentile of distances computed on clean data, permitting a $5\%$ false-positive rate on clean samples.
We use a combination of $K = 10$ transformations.
%


\vspace{0.05in}
\noindent
{\bf Image-based response generation.}
When a potential attack is detected, the LVLM is queried $K+1$ times independently: once 
with the original image and $K$ times with its transformed versions, 
$\{x, T_k(x)\}_{k=1}^K$. 
Each query uses the original instruction (captions in our case, although other query types are also possible).
This process yields a set of responses $R = \{r_k\}_{k=0}^{K}$.

\vspace{0.05in}
\noindent
{\bf Late detection.}
Given $R$, the set of textual responses generated for an image, the late detection phase further verifies the absence of an attack.
As before, the goal is to avoid unnecessarily invoking the costly external LLM consolidator during the final stage of the defense pipeline.
We note that this stage processes only those inputs that were not already deemed unattacked by the early detector. 

The central premise of the late detection phase is that discrepancies within the text embedding space serve as indicators of an attack.
A large discrepancy suggests that the generated responses diverge substantially from the expected semantic structure, leading to the sample being classified as attacked~\cite{zhang2023jailguard}. 
Furthermore, the response set for the remaining inputs is readily available, as it is required in the subsequent stage (response generation with LLM consolidator), eliminating the need for additional computation. 
Consequently, this phase is highly efficient while also providing an extra lightweight layer of verification.



This step can be implemented using various consistency-based detection methods, including embedding space distance measures \cite{reimers2019sentence}, question-answering checks over the content 
\cite{wu2024logical}, and LLM agreement judgments \cite{manakul2023selfcheckgpt, liu2023g}.

We adopt the adversarial detection mechanism proposed in JailGuard~\cite{zhang2023jailguard}, which analyzes the discrepancy between the response to the original input and the responses to its transformed variants.
This choice is motivated by its practicality and robustness in LLM settings. 
The method is model-agnostic, making it applicable to both open-and closed-source models,
and is lightweight and computationally efficient.


Briefly, each response $r_i$  is first mapped into a vector $V_i$ using a lightweight pretrained sentence embedding~\cite{reimers2019sentence}. 
A similarity matrix over the responses is constructed as 
\begin{equation} 
\label{eq:similarity} 
\small
S_{i,j} = \frac{V_i \cdot V_j}{|V_i| |V_j|}.
\end{equation}
Next, for each response $r_i$, we normalize the  $i^{th}$ row of the similarity matrix with respect to all responses.
Let $\|S_{i, \cdot}\|_1$ denotes the $\mathcal{L}_1$ norm of the $i^{th}$ row vector. 
This normalization yields a discrete similarity distribution $Q_i$:
\begin{equation}
\label{eq:probability}
\small
Q_i(k) = \frac{S_{i, k}}{\|S_{i, \cdot}\|_1}, \quad \text{for} \quad k \in \{1, \dots, K\}.
\end{equation}
$Q_i$ represents the semantic similarity of response $r_i$ to the full set $\mathcal{R}$.


The similarity matrix is converted to divergence matrix using KL divergence, quantifying the difference between any two similarity distributions.
The result is a matrix \(D\), in which each element \(D_{i,j}\) calculates the KL divergence between two distributions \((Q_i(x),Q_j(x))\),
\begin{equation}
\small
D_{i,j} = D(Q_i(x) \| Q_j(x)) = \sum_{x=1}^{K} Q_i(x) \log \left(\frac{Q_i(x)}{Q_j(x)}\right).
\end{equation}

If two responses \((r_i, r_j)\) exhibit significant semantic divergence, their corresponding entry \(D_{i,j}\) will be large. if any value in the divergence matrix \(D\) exceeds \(\tau_{late}\), the input is flagged as attacked and passed into the consolidation phase for final response generation. Otherwise, the original response $r_0$ is preserved as the final response. 

This step captures an additional $\sim$$1\%$ of clean images. 
The remaining false positives typically correspond to weak attacks that introduce only minor caption changes. For example, instead of “A man is standing in a field with two cows, one of which is brown and white$\dots$” the model produces “a man standing in a field with an apple in his hand, there’s a cow in the background$\dots$”



\vspace{0.05in}
\noindent
{\bf Response generation.}
%
Given the set of responses associated with an image 
$x$, the goal of this block is to generate a final output. 
For instance, when multiple captions are provided, it generates a single final caption. 
This block is applied only to responses corresponding to images deemed likely attacked during the late-detection stage (red arrows in \cref{fig:full_scheme_fig_v3});
responses from clean images (green arrows) bypass this block entirely. 
We emphasize that the response-generation block does not take the image as input, but operates solely on the textual descriptions.

The input text responses often differ from one another, as they are generated from transformed views,
as illustrated in \Cref{fig:teaser}. Moreover, they may exhibit various types of inconsistencies, including content discrepancies or residual artifacts remaining from the original attack. 
Therefore, the central challenge of this block is to consolidate all input responses into one coherent, unified response.


Our key idea is to treat these inconsistencies as informative signals: elements that consistently appear in 
$R$ (e.g., objects, attributes, relations) help guide the generation of a reliable final output;
Conversely,  elements appearing in only one or a few responses are likely incorrect.
For instance, in \Cref{fig:teaser}, “A red and white stop sign”, which appears in multiple responses, is highly likely to be correct, whereas “cars parked” ($r_1$) and “Give me a chance to see you” ($r_k$), which each appear in only a single response, are likely to be adversarial.
See \Cref{app:reasoning} for an additional reasoning example.

We propose using an external LLM for this purpose. Notably, this is the only stage that employs a large language model, and it is activated exclusively for inputs likely to be adversarial. Moreover, it operates solely on textual input rather than images, thereby reducing computational overhead.

Specifically, the LLM is prompted as follows (see \Cref{app:prompt} for the full prompt). First, it is informed that all responses originate from the same image, which may have been subjected to an adversarial attack. It is then instructed to

\noindent    (1) Identify the content of each response, focusing on the main objects.
    
\noindent    (2) Identify the consistent content across the responses, focusing on the main objects, attributes, scene settings, and object relationships.
   
\noindent  (3)   Evaluate consistency by tracking the frequency of each main object and the stability of its attributes across all responses.
 
\noindent (4)    Detect elements that appear only in the original response but seem inconsistent or unsupported, and disregard them.

\noindent (5)    Resolve conflicts by retaining the information confirmed in multiple responses and discarding inconsistent details.

\noindent (6) Generate a final response that integrates consistent content into a clear, natural description, excluding objects, attributes, or contextual details that appear only sporadically unless they are logically consistent with the rest.

Our response-generation stage differs in several ways from previous approaches. 
Sun et al.~\cite{sun2024safeguarding} select a final response by randomly sampling from the majority response group. In contrast, we synthesize all available responses to produce a unified output that reflects the full set.
Zhou et al.~\cite{zhou2024defending} also consolidate multiple responses. However, their approach requires access to the vision component (or the original image) during consolidation. By contrast, our method operates solely on the textual responses, resulting in greater efficiency and enabling the use of significantly smaller vision models.

\section{Evaluation}
\label{sec:evaluation}

This section evaluates the effectiveness and efficiency of our defense against attacks.
We test against adversarial examples generated by common baseline attacks, adhering to their standard setups for a fair comparison. 

\subsection{Experimental setup}

Image captioning offers a particularly strong defense evaluation, as it demands a deep and holistic understanding of the visual content. 
Unlike simple classification, captioning requires recognizing objects, attributes, spatial relationships, and scene context.
As a result, even subtle disruptions caused by adversarial or corrupted inputs become apparent in the generated descriptions. 


We evaluate our system against three representative and complementary attacks: MF-ii~\cite{zhao2023evaluating}, MixAttack~\cite{tu2023many}, and M-Attack~\cite{li2025frustratingly}, which are standard baselines in recent attack studies~\cite{xie2025chain, mei2025veattack}.
MF-ii and MixAttack introduce subtle perturbations that preserve visual appearance while degrading the model’s perceptual understanding.
M-Attack incorporates cropping-based transformations into its pipeline, yielding an attack that is robust to common defenses.


We follow the above attack protocols by generating adversarially manipulated images and feeding them directly into the target vision-language models. 
We conduct all evaluations using LLaVA-1.5-7B \cite{liu2024improved}, MiniGPT-v2 \cite{chen2023minigpt}, and Qwen3-VL-4B \cite{Qwen3-VL}, three compact yet widely adopted LVLMs. 
LLaVA and MiniGPT were evaluated in the MF-ii and MixAttack papers, while Qwen3-VL is a widely used SoTA model.
We focus on smaller architectures, whose accessibility and efficiency make them representative testbeds.

For response generation, we use the Gemma-3-27B model ~\cite{team2025gemma}, which is a moderately sized, open-source model.


We use clean images from MS-COCO~\cite{lin2014microsoft} and Flickr8k Expert ~\cite{hodosh2013framing} to ensure the availability of ground-truth captions for evaluation. 

For MS-COCO, the test set includes 1,000 images randomly sampled from the validation split, with target attack texts randomly drawn from the training set.
For Flickr8k Expert, the test set consists of 100 images randomly selected from the training split.



We evaluate the generated captions by comparing them to the ground truth captions. 
We use SentenceBERT (SBERT) \cite{reimers2019sentence} embedding space to compute the similarity between the generated text and the ground-truth caption. 
This metric provides semantically rich embedding space, with similarity scores that strongly correlate with human judgment of content equivalence without focusing on different phrasing.


\subsection{Results}

\noindent
{\bf Captioning evaluation.}
\Cref{tbl:main_cap} compares our training-free method with two variants of the training-free SmoothVLM defense: the original version using random masking~\cite{sun2024safeguarding} and our variant (“SVLM Ours”), which applies the same transformations as our method. 
We also report results with no defense applied (“Defense off”). 
Performance is evaluated on clean images and on images attacked by MF-ii, MixAttack, and M-attack.
Our method consistently outperforms all baselines; for example, under the MF-ii attack on LLaVA, it achieves a SentenceBERT score of 0.716, compared to 0.679 for~\cite{sun2024safeguarding} and 0.703 for our improved SmoothVLM variant. 
Finally, incorporating our transformation-selection strategy improves SmoothVLM’s performance.
%
We compare only with defenses applicable to small LVLMs (e.g., LLaVA, MiniGPTv2, and Qwen3-VL), excluding~\cite{zhou2024defending}, which relies on stronger models such as GPT-4o-Mini and Gemini.
We provide additional results on the Flickr8k Expert dataset in \Cref{app:dataset}.

\begin{table}[t]
\centering
\small
\begin{tabular}{ll|cccc}
\toprule
Model & Attack & Off  & \textsmaller{SVLM} & \textsmaller{SVLM} & \textsmaller{VALD} \\
 &  &   &  & Ours & (ours)  \\
\midrule

\multirow{4}{*}{LLaVA}
  & Clean & 0.722 & 0.713 & 0.722 & {\bf 0.723}  \\
  & MF-ii & 0.268 & 0.679 & 0.703 & {\bf 0.716} \\
  & MixAttk & 0.440 & 0.653 & 0.685 & {\bf 0.701}  \\
  & M-Attack & 0.360 & 0.667 & 0.654 & {\bf 0.673} \\
  \midrule
  
\multirow{4}{*}{MiniGPT}
  & Clean & 0.703 & 0.680 & 0.690 & {\bf 0.705}  \\
  & MF-ii & 0.217 & 0.663 & 0.674 & {\bf 0.706}  \\
  & MixAttk & 0.305 & 0.651 & 0.676 & {\bf 0.709}\\
  & M-Attack & 0.261 & 0.622 & 0.627 & {\bf 0.655} \\
\midrule

\multirow{4}{*}{Qwen3}
  & Clean & {\bf 0.723} & 0.702 & 0.718 & {\bf 0.723} \\
  & MF-ii & 0.534 & 0.667 & 0.700 & {\bf 0.711} \\
  & MixAttk & 0.565 & 0.660 & 0.695 & {\bf 0.704} \\
  & M-Attack & 0.484 & 0.668 & 0.680 & {\bf 0.689} \\
  
 \bottomrule
\end{tabular}
\caption{{\bf Captioning results on MS-COCO.}
VALD outperforms the baselines under all three attacks on three models according to the SentenceBERT score. 
}
\label{tbl:main_cap}
\end{table}



\noindent
{\bf Attack detection.}
\Cref{tbl:main_det} reports detection accuracy (the percentage of correctly classified inputs), recall (the fraction of adversarial images that are successfully detected), and precision (the fraction of inputs flagged as adversarial that are indeed adversarial). 
Overall, $93$-$96\%$ of clean images are correctly identified as clean, with nearly all of them filtered out by the fast early detector. Across all attacks and models, recall remains high, above 91\%
in every setting. The few adversarial images missed by the detector typically correspond to unsuccessful attacks that induce only minor caption changes. \Cref{app:detection} validate this observation by evaluating a setting in which the defense is always applied.

\begin{table}[t]
\centering
\small
\begin{tabular}{ll|ccc}
\toprule
Model & Attack & Acc & Precision & Recall \\
\midrule
\multirow{4}{*}{LLaVA}
  & Clean & 0.960 & 0 & N/A\\
  & MF-ii & 0.998 & 1 & 0.998\\
  & MixAttack & 0.991 & 1 & 0.991\\
  & M-Attack & 0.995 & 1 & 0.995 \\
\midrule

\multirow{4}{*}{MiniGPT}
  & Clean & 0.954 & 0 & N/A\\
  & MF-ii & 0.999 & 1 & 0.999\\
  & MixAttack & 0.998 & 1 & 0.998\\
  & M-Attack & 0.997 & 1 & 0.997  \\
\midrule

\multirow{4}{*}{Qwen3}
  & Clean & 0.930 & 0 & N/A \\
  & MF-ii & 0.987 & 1 & 0.987  \\
  & MixAttk & 0.912 & 1 & 0.912 \\
  & M-Attack & 0.955 & 1 & 0.955  \\

\bottomrule
\end{tabular}

\caption{{\bf Detection results.} 
The vast majority of clean images are correctly identified, with few adversarial images misclassified as clean images.}
\label{tbl:main_det}
\end{table}

\begin{figure*}[t]
    \centering
    \begin{tabular}{cc}
           \includegraphics[width=0.48\linewidth]{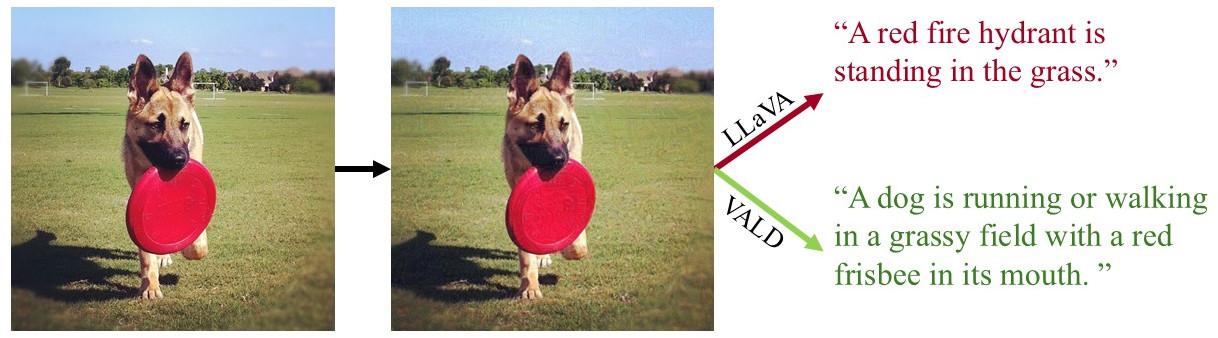} &
           \includegraphics[width=0.48\linewidth]{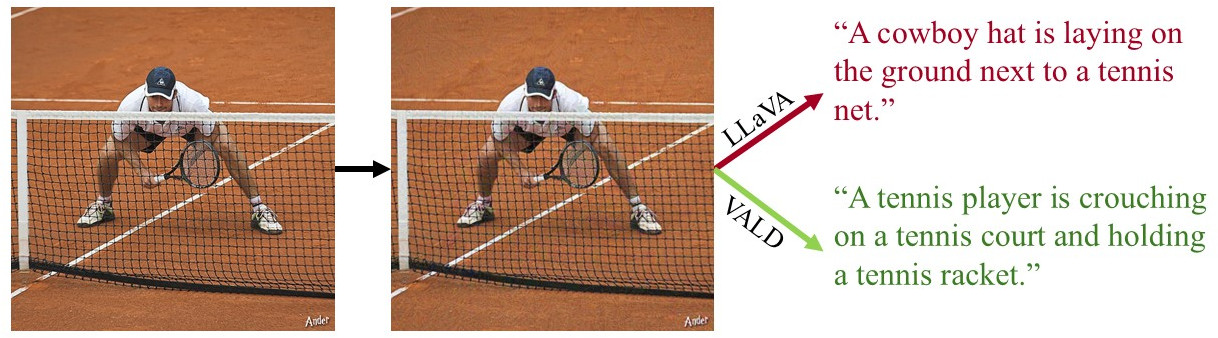} \\           
          \includegraphics[width=0.48\linewidth]{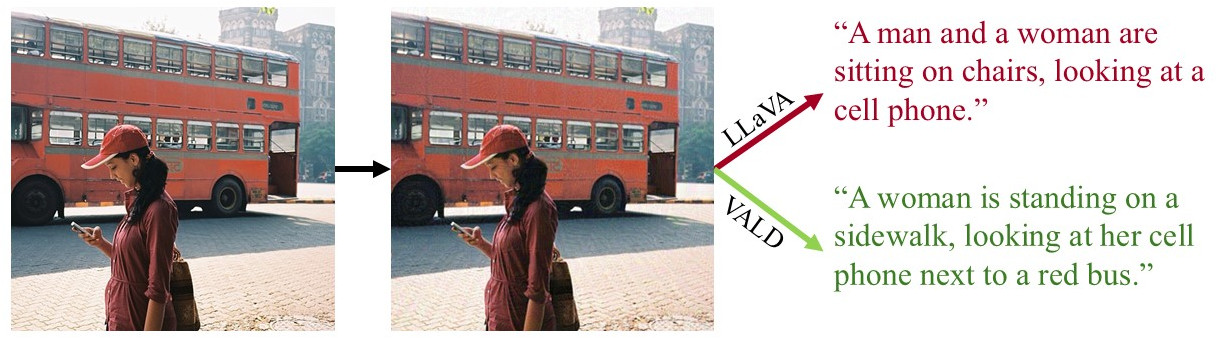} &
           \includegraphics[width=0.48\linewidth]{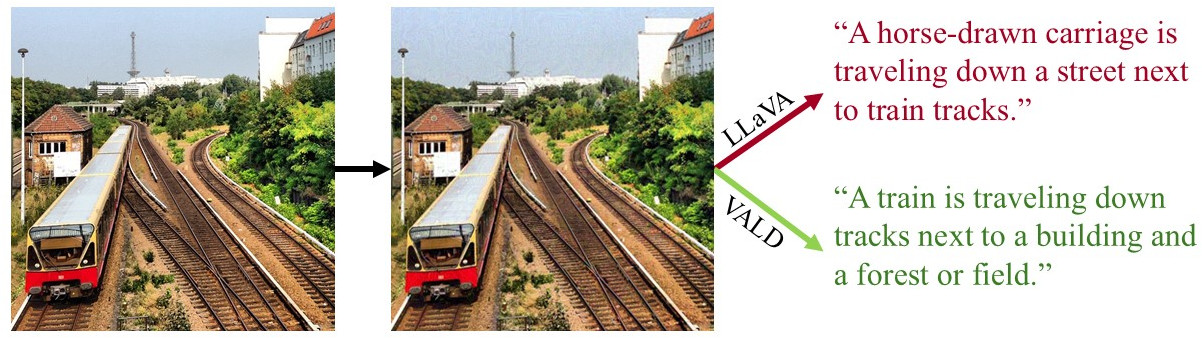} \\
            \includegraphics[width=0.48\linewidth]{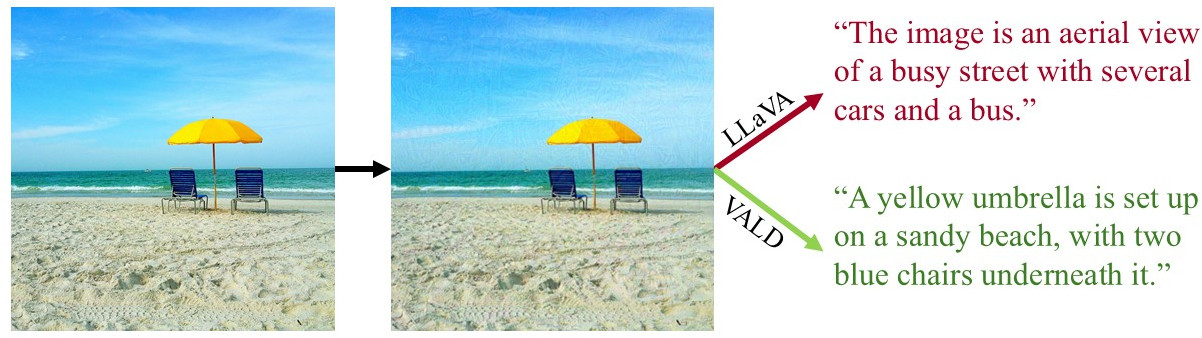} &

           \includegraphics[width=0.48\linewidth]{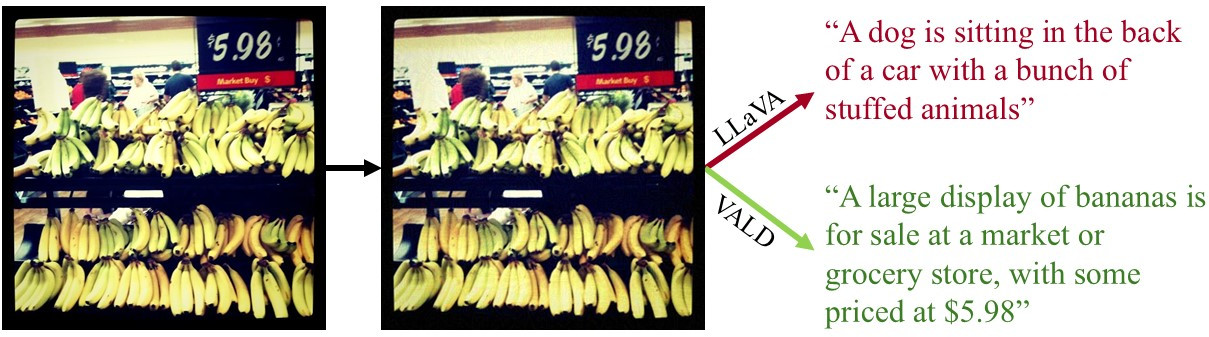} \\
      \hspace{-0.4in}    (a) input \hspace{0.1in} (b) attacked image  \hspace{0.1in} (c) captions &
     \hspace{-0.4in}     (a) input \hspace{0.1in} (b) attacked image  \hspace{0.1in} (c) captions
    \end{tabular}
    \caption{{\bf Qualitative results.}
    Although the attacked images (b) appear nearly identical to the originals (a), the LVLM incorrectly captions them (c). In contrast, our model successfully defends against the attack and generates accurate captions (c).
    The images were attacked via~\cite{zhao2023evaluating}.
    See \Cref{app:qualitativ2} for additional results.
    \label{fig:qualitative}}
\end{figure*}

\noindent
{\bf Comparison with training-based defense.}
\Cref{tbl:adv_training} compares our method with the adversarial training defense FARE~\cite{schlarmann2024robust}, which employs the same LLaVA model with a robust vision backbone.
VALD outperforms FARE across all attacks while maintaining clean performance.
Further analysis across noise levels appears in~\Cref{app:adv_training}.
Notably, under lower attack noise, FARE performs better (scores 0.617 - 0.671), yet VALD still surpasses it (scores 0.702 - 0.726).

\begin{table}[t]
\centering
\small
\begin{tabular}{l|c|c|c}
\toprule
Attack & Off & FARE & VALD \\
\midrule
Clean      & 0.722 & 0.672  & \textbf{0.723} \\
\midrule
MF-ii      & 0.268 & 0.570  & \textbf{0.716} \\
MixAttack  & 0.440 & 0.586  & \textbf{0.701} \\
M-Attack   & 0.360 & 0.517  & \textbf{0.673} \\
\bottomrule
\end{tabular}
\caption{{\bf Comparison with training-based defense. } VALD significantly outperforms FARE while maintaining clean performance.}
\label{tbl:adv_training}
\end{table}

\noindent
{\bf Qualitative evaluation.}
\Cref{fig:teaser,fig:qualitative} qualitatively compare our captions with an LVLM’s.
Although the attacked images are visually similar to the originals, the LVLM produces captions aligned with the targeted attack. In contrast, our method successfully resists the attack and generates accurate captions. For example, while our model correctly captions a dog with a frisbee, LLaVA misidentifies the image as a red fire hydrant.

\noindent
{\bf Runtime analysis.}
\Cref{tab:runtime} reports the best-case runtime (clean images) and worst-case runtime (adversarial images), and compares them to SmoothVLM.
Without any defense, caption generation takes $1.8$ seconds per image. 
When our defense is applied, the early detector requires $1.9$ seconds per image. 
It correctly identifies $\sim$$95\%$ of clean images, which are then excluded from the remainder of the pipeline. 
Images not classified as clean are forwarded to the LVLMs and the late detector, a stage that takes $18.2$ seconds. 
At this point, an additional $0.5\%$-$1\%$ of the remaining clean images are identified (depending on the model and the attack) and removed from further processing. 
Finally, response generation takes $2.8$ seconds.
By comparison, SmoothVLM processes every image and requires $52.7$ seconds per image.

\noindent
VALD runtime analysis was conducted using LLaVA model using on an NVIDIA RTX 6000 Ada GPU.
\Cref{app:runtime}
provides analysis of runtimes as a function of the percentage of clean images.
%




\subsection{Ablation study}



\noindent
{\bf Transformations parameters.}
\Cref{tbl:ablations_transformations_short} evaluates several content-preserving transformations, including large-scale random cropping, pixel masking, JPEG compression, and blurring.
We do not assume prior knowledge of the attack, and therefore use a mixture of transformations instead of a single type. This design improves our generality and robustness to adaptive attacks that target specific transformations (e.g., M-Attack with cropping). 
\Cref{app:trans_ablations} further analyzes performance as a function of the transformation parameters and $K$.

\begin{table}[t]
\centering
\small
\begin{tabular}{lcc}
\toprule
Method & Case & Runtime (s) \\
\midrule
VALD & Best & 1.9 \\
VALD & \bf{Worst} & 22.9 \\
\midrule
SmoothVLM & Always & 52.7 \\
\bottomrule
\end{tabular}
\caption{\textbf{Runtime (sec/image).} 
When clean images are detected by the early detection, VALD is 27× faster than SmoothVLM; when the entire defense is applied, it remains 2× faster. Notably, $\sim$$95\%$ of clean images are filtered at the early stage.}
\label{tab:runtime}
\end{table}

\begin{table}[t]
\centering
\small
\begin{tabular}{l|c|c|c|c|c}
\toprule
Attack & Crop & Masking & JPEG & Blur & Ours \\
\midrule
Clean      & 0.722 & \bf{0.726} & 0.723 & 0.723 & 0.723 \\
\midrule
MF-ii      & 0.716 & 0.694 & \bf{0.722} & 0.718  & 0.716 \\
MixAttack  & 0.697 & 0.672 & 0.707 &  \bf{0.716}  & 0.701\\
M-Attack   & 0.359 & 0.687 & 0.701 & \bf{0.707}   & 0.673 \\
\bottomrule
\end{tabular}
\caption{{\bf Transformations Ablations.} 
Our transformation mix improves robustness to (adaptive) attacks..}
\label{tbl:ablations_transformations_short}
\end{table}


\noindent
{\bf Perturbation magnitude.}
\Cref{app:epsilon_ablations} evaluates VALD across different perturbation budgets (i.e., the distance between clean and adversarial images~\cite{carlini2019evaluating}). 
As noise increases, undefended performance degrades, while applying the defense substantially improves robustness. 
Moreover, higher noise levels widen the performance gap between VALD and SmoothVLM. 

\section{Conclusion}
We presented a general, efficient, and training-free defense for Large Vision-Language Models against adversarial images. Our approach builds on two key ideas: a two-stage detection mechanism that efficiently filters most clean inputs, thereby avoiding costly defense procedures, and lightweight image transformations that expose adversarial behavior by consolidating multiple responses, where inconsistencies are weakened and consistent elements are reinforced. Extensive experiments demonstrate state-of-the-art accuracy while preserving performance on clean inputs. These results underscore the benefit of decoupling adversarial detection from expensive reasoning, enabling a practical and scalable defense for real-world LVLM deployment.


\section{Limitations}
\label{sec:limitation}

\label{sec:limitations}

Our defense is based on the assumption that genuine semantic content remains invariant under content-preserving transformations. 
However, when the same error appears consistently across multiple views, the method may converge on an incorrect yet mutually corroborated interpretation.
For instance, we observed an image containing two motorcycles (see \Cref{app:limitation}) to which substantial noise was added. Most transformed views hallucinated a third motorcycle, and the resulting caption incorrectly described three motorcycles.


In addition, there is an inherent trade-off between efficiency and robustness: as more images are classified as clean by the detector, the risk of failing to identify an attacked image increases.

\clearpage

\small
\bibliography{references}

\clearpage

\appendix
\normalsize

\section{Prompts}
\label{app:prompt}


In two stages of the method described in Section~\ref{sec:method}, we query the LVLM; below, we detail the corresponding prompts.

\noindent
{\bf Image-based response generation.}
Each LVLM is queried using the prompt {\em "provide a short description of the image"} for both the original and the transformed inputs. 

We also experimented with more targeted prompts for the transformed images, aiming to produce object level information needed by the consolidation stage (e.g., explicit listing of main objects). However, we found that they tended to produce shorter and less descriptive captions.


\noindent
{\bf Response generation.}
The full prompt used for consolidation of crops for the task of captioning is:

{\em "You are a multi-caption reasoning agent. You will receive multiple AI-generated image captions originated from the same image.
One is from the full image and others were generated with transformed versions of the original image as input.
The full image may have been affected by an adversarial attack and its caption should be treated with caution. The other transformed images might be adversarially affected as well, but less than the original image.

Your task is to consolidate these captions into a single, unified caption that accurately describes the content of the original image following a step-by-step procedure:

1. Identify the content of the captions. Focus on identifying clearly described main objects. Be precise and grounded in the text.

2. Identify consistent attributes and scene context. List attributes of the identified objects (descriptive features like color, size, appearance), the scene setting (environment, context), and any actions or relationships described in the caption. Do not speculate. If a detail is implied but not stated, omit it.

3. Evaluate consistency: Keep track of how frequently each object appears across the original image caption and the transformed image captions. Objects consistently appearing in most of the captions from the transformed images are highly likely to be present in the original image. Give higher weight to object presence confirmed across multiple captions, especially if the original caption significantly diverges from the majority.

4. Check caption alignment: Detect and ignore elements that only appear in the original caption if they seem inconsistent or unsupported.

5. Resolve conflicts using a majority-consensus rule: Prefer information that appears in multiple captions. Identify reliable objects and discard any that are not corroborated by multiple captions. Group semantically equivalent terms (e.g., "car" and "vehicle").
    
Your output:

1. A final caption that combines consistent content into a clear and natural description: In the final caption avoid including objects, attributes, or context mentioned only in one caption unless clearly confirmed by other captions or logically consistent.

2. An explanation of the data consolidation process: Make sure not to include parts of the explanation in the final generated caption!
        
The original image caption is: <original caption>.

The transformed images captions are: < transformations captions>"
}

\begin{figure}[t]
    \centering
    \includegraphics[width=0.55\columnwidth]{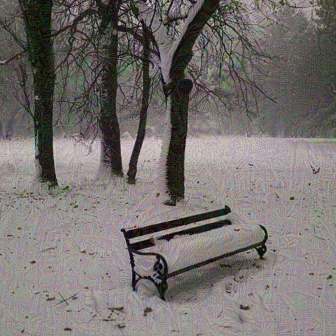}
    \caption{{\bf Consolidation process explanation.} 
 The following elements are mentioned by most captions and form the basis of the final consolidated description: a snow-covered park or forest, a wooden bench, trees, and the bench being covered in snow.
Conversely, the following details either conflict with the majority or appear only sporadically; therefore, they are treated as inconsistent and omitted from the final caption: a coconut beauty product, a bird on the bench, a green object on the tree, and a black-and-white filter.
    }
    \label{fig:app_cot}
\end{figure}

\section{ Example For Reasoning Process}
\label{app:reasoning}

\begin{figure*}[t]
    \centering
    \begin{tabular}{cc}
     \includegraphics[width=0.48\linewidth]{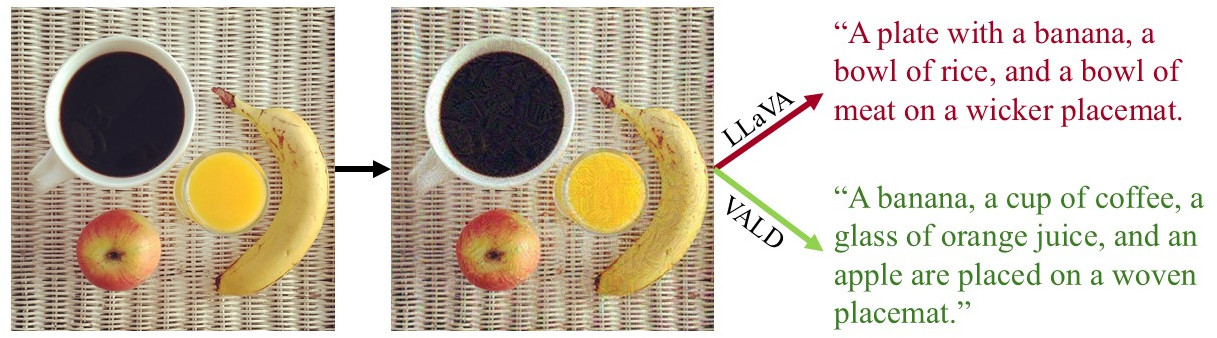} &
     \includegraphics[width=0.48\linewidth]{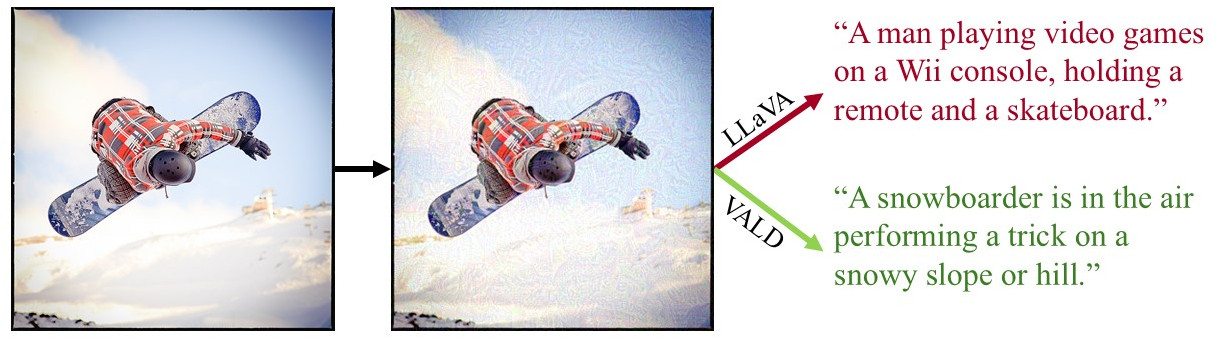}  \\
      \includegraphics[width=0.48\linewidth]{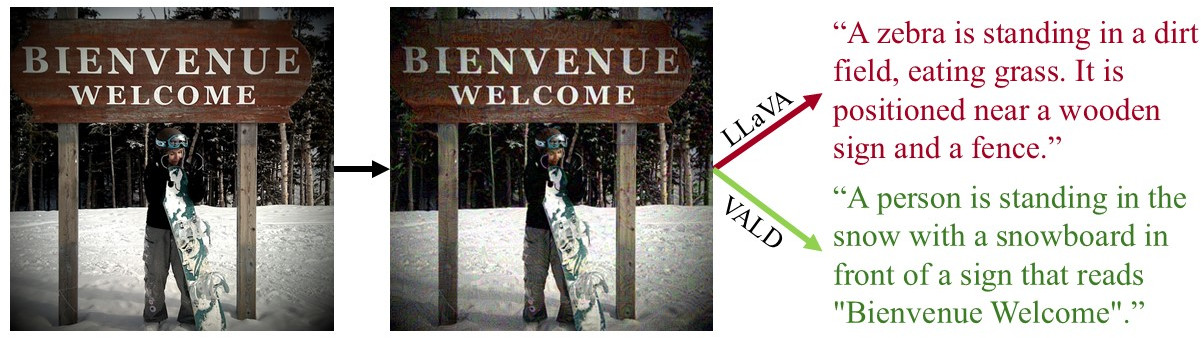} &
      \includegraphics[width=0.48\linewidth]{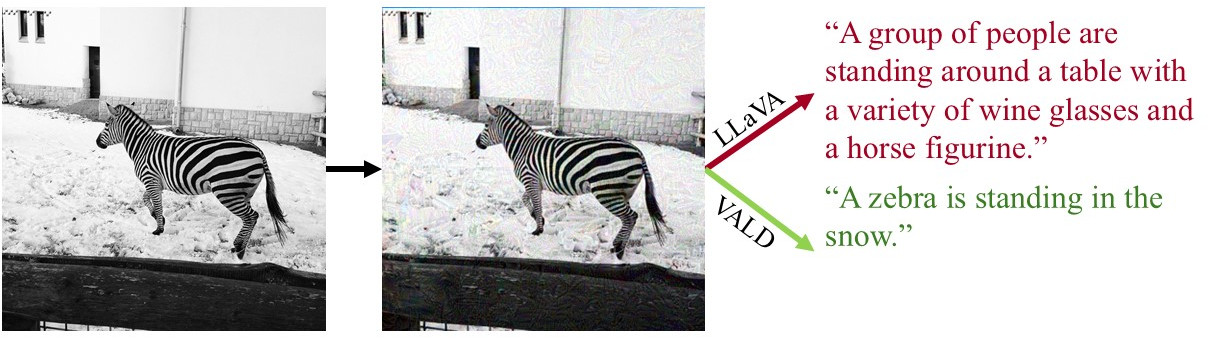} \\
     \includegraphics[width=0.48\linewidth]{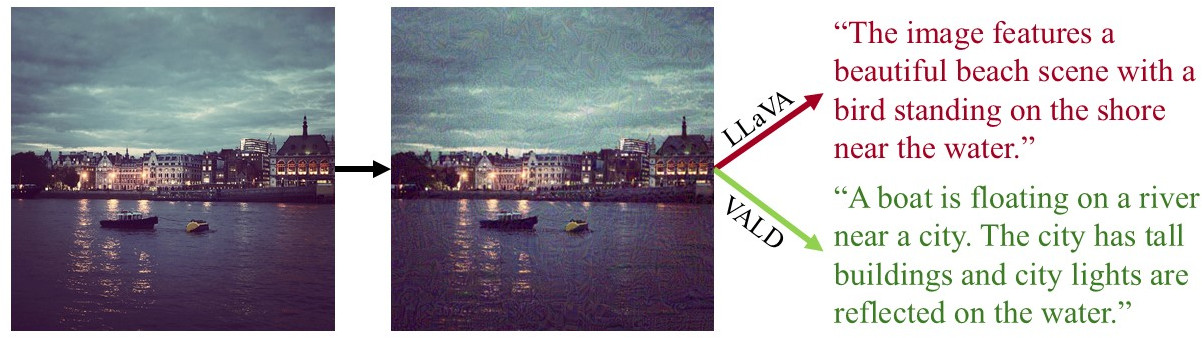} &
       \includegraphics[width=0.48\linewidth]{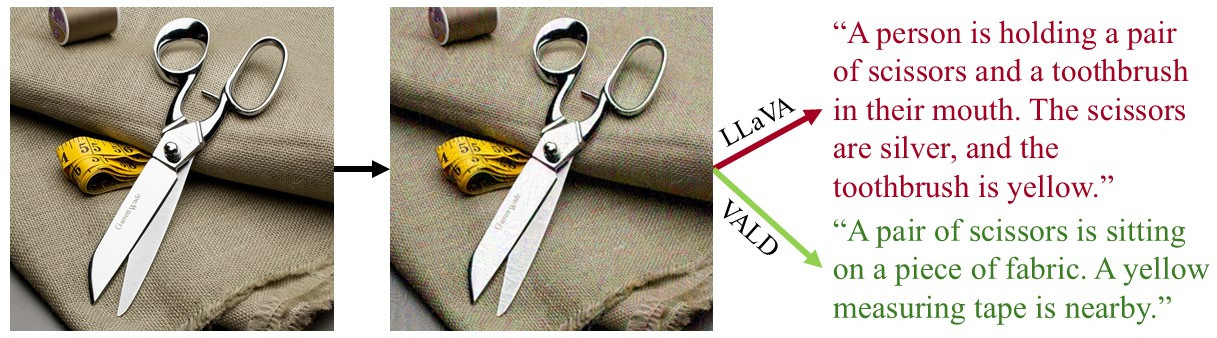} \\

      \hspace{-0.4in}    (a) input \hspace{0.1in} (b) attacked image  \hspace{0.1in} (c) captions &
     \hspace{-0.4in}     (a) input \hspace{0.1in} (b) attacked image  \hspace{0.1in} (c) captions
    \end{tabular}
    \caption{{\bf Qualitative results.}
    Although the attacked images (b) appear nearly identical to the originals (a), the LVLM incorrectly captions them (c). In contrast, our model successfully defends against the attack and generates accurate captions (c).
    The images were attacked via~\cite{zhao2023evaluating}.
    \label{fig:qualitative2}}
\end{figure*}

We next present an example of the consolidation process in the response generation stage. Given the input image in \Cref{fig:app_cot}, the captions generated from the transformed views include the following (out of a total of $10$ captions):
\begin{enumerate}
    \item 
    "The image features a wooden park bench sitting in the snow, surrounded by trees. The bench appears to $\dots$ The scene is captured in black and white, adding a sense of nostalgia to the image."
    \item 
    "The image features a snowy park with a wooden bench sitting in the snow. The bench is situated near a tree, and there is a bird perched on the back of the bench $\dots$"
    \item
    "The image features a snow-covered park with a wooden bench sitting in the middle of the snow. The bench is positioned under a tree, $\dots$ The tree is adorned with a green object, possibly a decoration or a piece of clothing."
\end{enumerate}

In the explanation of the data consolidation process used to produce the output (Section~\ref{app:prompt}), we obtain the following explanation:

{\em "The original caption describes an image of a coconut beauty product, which is inconsistent with the crop captions. The crop captions consistently describe a snow-covered park or forest with a wooden bench and trees. The bench being covered in snow is also a consistent detail. 
Details like a bird on the bench, a green object on the tree, or the black and white filter appear in only a few captions and are therefore omitted. 
The final caption consolidates the most frequently mentioned elements - the bench, snow, trees, and the park/forest setting - to provide an accurate description of the image based on the majority of the provided evidence."}

\section{Additional Qualitative Results}
\label{app:qualitativ2}

\Cref{fig:qualitative2} provides additional qualitative results, in addition to those presented in \Cref{fig:qualitative}.
It verifies that, despite minimal visual differences, the LVLM yields attack-aligned captions, while our method resists the attack and produces correct outputs.

\section{Visualization Of The Limitation}
\label{app:limitation}

\begin{figure}[t]
    \centering
    \includegraphics[width=0.55\columnwidth]{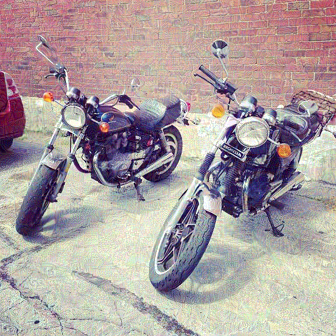}
    \caption{{\bf Example of a failure case in the consolidation process.} 
    Most transformed views hallucinate a third motorcycle, causing the consolidated caption to mistake.}
    \label{fig:consolidation_failure}
\end{figure}

\Cref{sec:limitations} describes a failure case in which the input image contains two motorcycles parked side by side, yet some transformed views incorrectly report three motorcycles. 
\Cref{fig:consolidation_failure} illustrates this input image.
The final consolidated response is:
{\em "The image features three motorcycles parked in a row on a concrete or cement surface. The motorcycles are of different sizes and positioned next to each other, near a brick wall."} 

The explanation of the consolidation process for this caption is as follows:
{\em "The consolidation process began by identifying the core objects present in all captions: motorcycles. The number of motorcycles varied between 'two' and 'three'. 'three' appeared more frequently across the crop captions and is therefore prioritized. Mentions of two motorcycles may be explained by cropping effects."}

This case may be explained by the presence of visible car edges on the left side of the image, which some transformed captions may have misinterpreted as parts of an additional motorcycle.

\section{Additional Detection Analysis}
\label{app:detection}

\begin{table}[t]
\centering
\small
\begin{tabular}{ll|ccc}
\toprule
Model & Attack & Off  &  \textsmaller{VALD} & Always On  \\
 
\midrule

\multirow{4}{*}{LLaVA}
  & Clean & 0.722 &  0.723 & \bf{0.733}  \\
  & MF-ii & 0.268 & \bf{0.716} & \bf{0.716} \\
  & MixAttk & 0.440 & \bf{0.701} & \bf{0.701}  \\
  & M-Attack & 0.360 & \bf{0.673} & \bf{0.673}\\
  \midrule
  
\multirow{4}{*}{MiniGPT}
  & Clean & 0.703 & 0.705 & \bf{0.726} \\
  & MF-ii & 0.217 & \bf{0.706} & \bf{0.706} \\
  & MixAttk & 0.305 & \bf{0.709} & \bf{0.709}\\
  & M-Attack & 0.261 & \bf{0.655} & \bf{0.655} \\
\midrule

\multirow{4}{*}{Qwen3}
  & Clean & 0.723 & 0.723 & \bf{0.725} \\
  & MF-ii & 0.534 & 0.711 & \bf{0.712} \\
  & MixAttk & 0.565 & 0.704 & \bf{0.705} \\
  & M-Attack & 0.484 & 0.689 & \bf{0.691} \\
  
 \bottomrule
\end{tabular}
\caption{{\bf Defense performance with and without detection.}
Enabling detection preserves VALD defense performance, yielding results nearly identical to always applying the defense.
}
\label{tbl:cap_def_all_on}
\end{table}

\Cref{tbl:cap_def_all_on} reports defense performance both with detection enabled and with the defense always applied. Across all attacks and models, the results are nearly identical, indicating that the detection stage has little effect on overall defense effectiveness while substantially reducing runtime. For Qwen3-VL, where detection performance is somewhat lower than for LLaVA and MiniGPT, the near-identical defended results suggest that the missed adversarial images correspond to ineffective attacks and are therefore classified as clean by the detection mechanism.

\begin{table}[t]
\centering
\small
\begin{tabular}{c|l|c|c|c}
\toprule
$\epsilon$ & Attack & Off & FARE & VALD \\
\midrule
 & Clean      & 0.722 & 0.672 & \textbf{0.723} \\
 \midrule
 \multirow{3}{*}{8}
 & MF-ii      & 0.282  & 0.617 & \textbf{0.726} \\
 & MixAttack  & 0.603  & 0.644 & \textbf{0.720} \\
 & M-Attack   & 0.429  & 0.671 & \textbf{0.702} \\
\midrule

\multirow{3}{*}{16}
 & MF-ii      & 0.268 & 0.570 & \textbf{0.716} \\
 & MixAttack  & 0.440 & 0.586 & \textbf{0.701} \\
 & M-Attack   & 0.360 & 0.517 & \textbf{0.673} \\

\bottomrule
\end{tabular}
\caption{\textbf{Robustness under different perturbation budget ($\epsilon$).}
VALD consistently outperforms the adversarially trained baseline FARE, with a larger performance gap at higher perturbation levels.
}

\label{tbl:adv_training_mul_eps}
\end{table}


\section{Adversarial Training Analysis}
\label{app:adv_training}

Table~\ref{tbl:adv_training_mul_eps} compares VALD with the robust FARE model under different perturbation budgets $\epsilon$. In our evaluation, we use FARE's LLaVA-based variant, where the original LLaVA vision backbone is replaced with the  FARE vision backbone. 
We evaluate the strongest available version of FARE, which is trained with adversarial examples at $\epsilon=4$. At $\epsilon=8$, VALD achieves higher robustness than FARE across all attack types. When the attack strength increases to $\epsilon=16$, the gap becomes larger. For example, under MF-ii with $\epsilon=8$ VALD achieves 0.726 compared to 0.617 for FARE, and for $\epsilon=16$ VALD achieves 0.716 compared to 0.570 for FARE.

\begin{table}[t]
\centering
\small
\begin{tabular}{lc}
\toprule
Stage  & \shortstack{Runtime \\ (Average)}  \\
\midrule

Caption generation (Defense off) & 1.8 \\
\midrule

Caption generation with ED \textbf{(Best Case)} & 1.9 \\

LVLM querying & 18.0 \\
Late Detector & 0.2 \\

Response generation & 2.8 \\

\midrule
\textbf{VALD total runtime (Worst Case)} & 22.9 \\
\textbf{SmoothVLM total runtime} & 52.7 \\

\bottomrule
\end{tabular}

\caption{{\bf Runtime breakdown (sec / image).} 
Images filtered by the Early Detector (ED) correspond to the best case runtime, while the worst case occurs when images pass through the full pipeline.}

\label{tab:runtime_breakdown}
\end{table}

\begin{table}[t]
\centering
\small
\begin{tabular}{lccccc}
\toprule
Attack & Clean \% & Acc & Rec & Prec & Time \\
\midrule
MF-ii & 50 & 0.979 & 0.998 & 0.962 & 12.8 \\
MF-ii & 95 & 0.962 & 1 & 0.568 & 3.8 \\
\midrule
MixAttk & 50 & 0.973 & 0.992 & 0.956 & 12.9 \\
MixAttk & 95 & 0.961 & 1 & 0.562 & 3.8 \\
\midrule
M-Attack & 50 & 0.979 & 0.994 & 0.965 & 12.8 \\
M-Attack & 95 & 0.962 & 1 & 0.568 & 3.8 \\
\bottomrule
\end{tabular}
\caption{\textbf{Runtime as a function of clean images percentage.}
Average runtime decreases as the proportion of clean images increases.
}

\label{tab:vald_detection}
\end{table}


\section{Additional Runtime Analysis}
\label{app:runtime}

\textbf{Runtime breakdown.} 
Table~\ref{tab:runtime_breakdown} provides a detailed breakdown of the runtime across the different stages of the VALD pipeline. Images filtered by the Early Detector correspond to the best-case runtime. If an image is filtered by the Late Detector, the runtime before the response generation stage is approximately 20.1 seconds. 
All runtime measurements were performed on VALD LLaVA setup and were conducted on a NVIDIA RTX 6000 Ada GPU.

\noindent
\textbf{Runtime as a function of clean images percentage.} 
Table~\ref{tab:vald_detection} reports the average runtime of VALD (per image) for different proportions of clean images. 
In realistic settings, most inputs are clean. Therefore we evaluate two scenarios, one in which 50\% of the images are clean and another in which 95\% are clean. 
Since the early detection identifies approximately 95\% of clean images, these filtered inputs incur a runtime that is nearly identical to that of running without defense (1.8s vs.\ 1.9s).
As a result, the average runtime decreases as the proportion of clean images increases.

In both scenarios, recall remains above 99\%, indicating that almost no adversarial images are misclassified as clean. Overall accuracy also stays high across both scenarios. As expected, precision decreases when the proportion of clean images increases. This is because the precision measures how many of the samples flagged as adversarial are truly adversarial. When clean images are much more common, even a small number of clean images misclassified as adversarial constitutes a larger fraction of the samples classified as adversarial.

\begin{table*}[t]
\centering
\small
\begin{tabular}{ll|cccc|ccc}
\toprule
Model & Attack & Off  & \textsmaller{SVLM} & \textsmaller{SVLM} & \textsmaller{VALD} & Acc & Prec & Rec \\
 &  &   &  & Ours & (ours)  \\
\midrule
\multirow{4}{*}{LLaVA}
  & Clean    & 0.713 & 0.690 & 0.704 & \textbf{0.722} & 0.55 & 0 & N/A \\
  & MF-ii    & 0.213 & 0.642 & 0.694 & \textbf{0.710} & 1 & 1 & 1\\
  & MixAttk  & 0.382 & 0.608 & 0.643 & \textbf{0.664} & 1 & 1 & 1\\
  & M-Attack & 0.347 & 0.651 & 0.652 & \textbf{0.663} & 0.98 & 1 & 0.98\\
\bottomrule
\end{tabular}
\caption{\textbf{Captioning and detection results on the Flickr8k Expert dataset.}
VALD surpasses the SVLM baseline in captioning performance while correctly detecting nearly all adversarial images.}
\label{tbl:main_cap_flickr}
\end{table*}

\setcounter{table}{11}
\begin{table*}[t]
\centering
\small
\begin{tabular}{l|ccc|ccc|ccc|ccc}
\toprule
  & \multicolumn{3}{c|}{Crop} 
 & \multicolumn{3}{c|}{Pixel masking}
 & \multicolumn{3}{c|}{JPEG}
 & \multicolumn{3}{c}{Blur} \\
\cmidrule(lr){2-4} \cmidrule(lr){5-7} \cmidrule(lr){8-10} \cmidrule(lr){11-13}
 Attack & 95\% & 90\% & 85\% & 10\% & 15\% & 20\% & 30 & 40 & 50 & 5 & 7 & 9 \\
\midrule
  Clean & 0.722 & 0.722 & 0.722 & 0.726 & 0.725 & 0.724 & 0.723 & 0.722 & 0.722 & 0.722 & 0.722 & 0.723 \\
  MF-ii & 0.716 & 0.716 & 0.713 & 0.694 & 0.693 & 0.689 & 0.722 & 0.718 & 0.712 & 0.700 & 0.718 & 0.718 \\
  MixAttack & 0.697 & 0.693 & 0.688 & 0.672 & 0.677 & 0.676 & 0.707 & 0.696 & 0.675 & 0.677 & 0.713 & 0.716 \\
  M-Attack & 0.359 & 0.359 & 0.360 & 0.687 & 0.686 & 0.680 & 0.701 & 0.679 & 0.653 & 0.600 & 0.697 & 0.707 \\
\bottomrule
\end{tabular}
\caption{\textbf{Transformation parameter ablation.}
For cropping and pixel masking, performance is best when most of the original image is preserved, whereas for JPEG compression and blur, stronger transformations provide better robustness.}
\label{tbl:ablations_transformations_full}
\end{table*}
\setcounter{table}{10}
\begin{table}[t]
\centering
\small
\begin{tabular}{ll|cccc}
\toprule
Model &  $\epsilon$ & Off  & \textsmaller{SVLM} & \textsmaller{SVLM} & \textsmaller{VALD} \\
 &  &   &  & Ours & (ours)  \\
\midrule
Clean & --   & 0.722 & 0.713 & 0.722 & \textbf{0.723} \\
\cmidrule(lr){1-6}
\multirow{3}{*}{MF-ii}
  & $8$  & 0.282 & 0.684 & 0.712 & \textbf{0.726} \\
  & $16$ & 0.268 & 0.679 & 0.703 & \textbf{0.716} \\
  & $24$ & 0.262 & 0.665 & 0.695 & \textbf{0.704} \\
\cmidrule(lr){1-6}
\multirow{3}{*}{MixAttack}
  & $8$  & 0.603 & 0.683 & 0.704 & \textbf{0.720} \\
  & $16$ & 0.440 & 0.653 & 0.685 & \textbf{0.701} \\
  & $24$ & 0.346 & 0.616 & 0.662 & \textbf{0.673} \\
  \cmidrule(lr){1-6}
\multirow{3}{*}{M-Attack}
  & $8$  & 0.429 & 0.681 & 0.682 & \textbf{0.702} \\
  & $16$ & 0.360 & 0.667 & 0.654 & \textbf{0.673} \\
  & $24$ & 0.341 & \textbf{0.648} & 0.637 & 0.623 \\
\bottomrule
\end{tabular}
\caption{{\bf Impact of perturbation.} 
Increasing noise degrades performance without defense (Off), while our defense improves robustness and increasingly outperforms SmoothVLM (SVLM).}
\label{tbl:ablations_epsilon}
\end{table}



\section{Evaluation On An Additional Dataset}
\label{app:dataset}

\textbf{Flickr8k Expert evaluation.}
Table~\ref{tbl:main_cap_flickr} presents results on the Flickr8k Expert dataset ~\cite{hodosh2013framing}, which contains high quality captions. In this dataset, each image caption pair is evaluated by three expert annotators with scores ranging from 1 to 4, where a score of 1 indicates that the caption does not correspond to the image and a score of 4 indicates a fully correct description of the visual content. For our evaluation we randomly select 100 images with high expert scores, which serve as ground truth for evaluating the generated captions.

Consistent with the main paper results, our method outperforms the SVLM baseline across all attack settings, showing that its advantage is not dataset specific. 

While our detection mechanism does not rely on attack-specific behavior, it does depend on modeling the distribution of clean images behavior and therefore requires threshold calibration for different data distributions.
Using the thresholds calibrated on COCO, we find that nearly all adversarial images are still correctly identified as adversarial, but more clean images are incorrectly flagged as adversarial. In our framework, such errors mainly increase computation, since misclassifying a clean image as adversarial only causes the defense to be applied unnecessarily and does not harm captioning performance.

\section{Perturbation Budget Ablations}
\label{app:epsilon_ablations}

\noindent
\Cref{tbl:ablations_epsilon} evaluates defense performance across different perturbation budgets ($\epsilon$).  
As $\epsilon$ increases, undefended performance degrades, while applying the defense improves robustness. 
Additionally, the gap between VALD and SmoothVLM increases at higher noise levels, For example, at $\epsilon = 16$, VALD outperforms SmoothVLM by 4\%, and at $\epsilon = 24$ the gap further increases to 5\% under MixAttack.

\section{Additional Transformations Ablations}
\label{app:trans_ablations}
\subsection{Transformations Parameters Ablations}
Table~\ref{tbl:ablations_transformations_short} reports the best performing parameter configuration for each transformation used in our defense.
Table~\ref{tbl:ablations_transformations_full} provides the full ablation across the evaluated parameter ranges to analyze the sensitivity of the method to these choices.
Cropping percentages indicate the retained spatial extent along both image axes; pixel masking percentages correspond to the fraction of pixels randomly suppressed in the image; JPEG values denote the compression quality parameter, where lower values imply stronger compression; blur values correspond to the size of the applied gaussian kernel.
The results reveal two clear trends. For cropping and pixel masking, better performance is achieved when most of the original image is preserved (e.g., $95\%$ crop and $10\%$ masking). In contrast, for JPEG compression and blur, stronger transformations lead to better robustness. As a future work, we plan to explore combinations of transformations, where each randomized image undergoes several transformations, making it harder for adversarial perturbations to remain effective.


\subsection{Ablation On \(K\)}

Table~\ref{tbl:ablations_k} examines the effect of the number of transformations \(K\).

The impact of \(K\) varies across attacks. For MF-ii and MixAttack, the applied transformations are relatively effective as a defense, so even \(K=6\) transformed views is sufficient for reliable consolidation, and performance remains largely stable as \(K\) increases.
In contrast, M-Attack is more resilient to transformations (especially cropping), and therefore benefits more from increasing \(K\),  improving from \(0.532\) at \(K=4\) to \(0.673\) at \(K=10\). 
However, this improvement begins to diminish beyond \(K\approx\!\!10\).
As noted in the limitations section, when the consolidating LLM receives too many captions with consistent errors, these errors may no longer be treated as outliers. 
Therefore, increasing \(K\) indefinitely not only raises computational cost, but can also eventually degrade performance.
Overall, these results suggest that a moderate number of transformations captures most of the robustness gains, while maintaining a reasonable trade-off with computational overhead.

\section{SmoothVLM Implementation Details}
\label{app:smoothvlm}
SmoothVLM~\cite{sun2024safeguarding} operates as follows. The LVLM is queried on both the adversarial image and its transformed versions, where the original method uses random pixel masking.
The caption generated from the adversarial image is then compared with the captions generated from the transformed images, and an external LLM judge determines whether each pair is synonymous. The captions are partitioned according to this classification, and the final caption is randomly selected from the majority group.

To ensure fairness, we evaluate SmoothVLM in a setting that closely matches our own. For each comparison, we use the same LVLM backbone as in our method, as indicated in the tables (e.g., LLaVA, MiniGPT, and Qwen3). We also use \textbf{Gemma-27B} as the external LLM judge, matching the model used in our consolidation stage, and set the number of transformations to $K=10$ for both methods. 
In addition to the original SmoothVLM baseline (SVLM), we report a variant denoted SVLM-Ours, in which SmoothVLM uses the same transformation set as our method - JPEG compression, blur, random cropping, and random masking. This ensures an identical experimental setup and enables a more direct comparison between the two methods.


\setcounter{table}{12}
\begin{center}
\small
\begin{tabular}{l|cccccc}
\toprule
Attack & \multicolumn{5}{c}{Number of Transformations $K$} \\
\cmidrule(lr){2-7}
 & 2 & 4 & 6 & 8 & 10 & 12 \\
\midrule
  MF-ii     & 0.689 & 0.709 & 0.716 & 0.714 & 0.716 & \bf{0.717} \\
  MixAttack & 0.664 & 0.694 & \bf{0.703} & 0.700 & 0.701 & 0.702 \\
  M-Attack  & 0.398 & 0.532 & 0.608 & 0.649 & \bf{0.673} & 0.640 \\
\bottomrule
\end{tabular}
\captionof{table}{\textbf{Effect of the number of transformations $K$.} A moderate number of transformed views captures most of the robustness gains, while larger $K$ saturate and possibly reduce performance.}
\label{tbl:ablations_k}
\end{center}

\end{document}